\documentclass[10pt,twocolumn,letterpaper]{article}

\usepackage{theme}

\renewcommand{\paragraph}[1]{\vspace{.5em}\noindent\textbf{#1.}}

\usepackage{booktabs}
\usepackage{multirow}
\usepackage{siunitx}
\usepackage{makecell} \usepackage{booktabs}
\usepackage{array}

\definecolor{cvprblue}{rgb}{0.21,0.49,0.74}
\usepackage[pagebackref,breaklinks,colorlinks,allcolors=cvprblue]{hyperref}

\title{Refining Visual Artifacts in Diffusion Models via Explainable AI-based Flaw Activation Maps}

\author{
Seoyeon Lee\textsuperscript{*} \quad
Gwangyeol Yu\textsuperscript{*} \quad
Chaewon Kim\textsuperscript{*} \quad
Jonghyuk Park\textsuperscript{$\dagger$}\\
Kookmin University\\
\centerline{{\tt\small \{tjdus0223, rhkdduf627, kimcwbf, jonghyuk\}@kookmin.ac.kr}}
}

\begin{document}
\maketitle
\begingroup
\renewcommand\thefootnote{}
\footnotetext{* Equal contribution.}
\footnotetext{$\dagger$ Corresponding author.}
\endgroup

\begin{abstract}
Diffusion models have achieved remarkable success in image synthesis. However, addressing artifacts and unrealistic regions remains a critical challenge. We propose self-refining diffusion, a novel framework that enhances image generation quality by detecting these flaws. The framework employs an explainable artificial intelligence (XAI)-based flaw highlighter to produce flaw activation maps (FAMs) that identify artifacts and unrealistic regions. These FAMs improve reconstruction quality by amplifying noise in flawed regions during the forward process and by focusing on these regions during the reverse process. The proposed approach achieves up to a 27.3\% improvement in Fréchet inception distance across various diffusion-based models, demonstrating consistently strong performance on diverse datasets. It also shows robust effectiveness across different tasks, including image generation, text-to-image generation, and inpainting. These results demonstrate that explainable AI techniques can extend beyond interpretability to actively contribute to image refinement. The proposed framework offers a versatile and effective approach applicable to various diffusion models and tasks, significantly advancing the field of image synthesis.
\end{abstract}
\section{Introduction}
\label{sec:intro}

In recent years, generative models have made remarkable progress, with diffusion models emerging as a powerful framework for various computer vision tasks such as image synthesis \cite{ho2020ddpm, song2020score}. These models have demonstrated a superior performance over conventional generative adversarial networks (GANs) \cite{goodfellow2014gans, brock2019large, karras2019stylegan} and variational autoencoders (VAEs) \cite{kingma2014vae}, achieving state-of-the-art results in generating high-quality and diverse images.

However, despite their impressive generative capabilities, diffusion models have several inherent limitations. Notably, they can produce artifacts, subtle inconsistencies, and unrealistic details in generated images, which negatively affect the overall quality and realism of the output \cite{aithal2024hallucinations}. Identifying and mitigating these specific generation-related flaws remains a persistent challenge.

\begin{figure}[t]
  \centering
  \includegraphics[width=\linewidth]{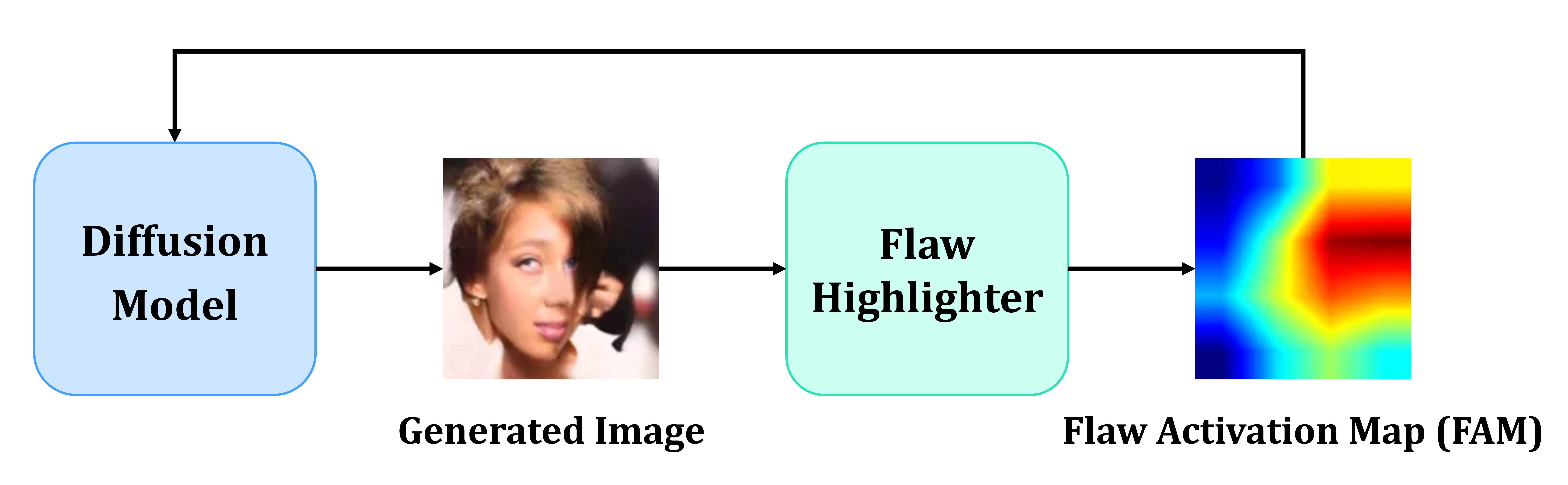}
  \caption{Overview of self-refining diffusion framework.}
  \label{fig:overview}
\end{figure}

To overcome these challenges, enabling generative models to recognize and understand the flaws in their outputs is crucial. In this context, explainable artificial intelligence (XAI) techniques provide valuable insights into model behavior, helping to interpret the decision-making process within these models \cite{samek2019xai}. Although some studies have explored the application of XAI techniques to diffusion models, they have primarily focused on interpretability rather than performance enhancement \cite{longo2021xai}.

In this study, we propose “Self-Refining Diffusion,” a novel framework that improves the image quality of diffusion models through an XAI-based flaw detection mechanism. As illustrated in \cref{fig:overview}, the proposed framework operates as a unique feedback loop that identifies and resolves flaws such as artifacts and subtle inconsistencies during training.

The proposed framework incorporates a flaw highlighter to detect flawed regions in generated images. Trained to distinguish between real and generated images, the highlighter uses class activation mapping (CAM) techniques \cite{zhou2016cam,selvaraju2017gradcam} to identify flawed regions, that is, areas of an image that most strongly influence its decision that the image is fake. The identified regions are then incorporated into the diffusion process in two ways: first, as noise components, and second, within the attention mechanisms of the denoising network in the diffusion model.

\begin{figure*}[t]
  \centering
  \includegraphics[width=\textwidth]{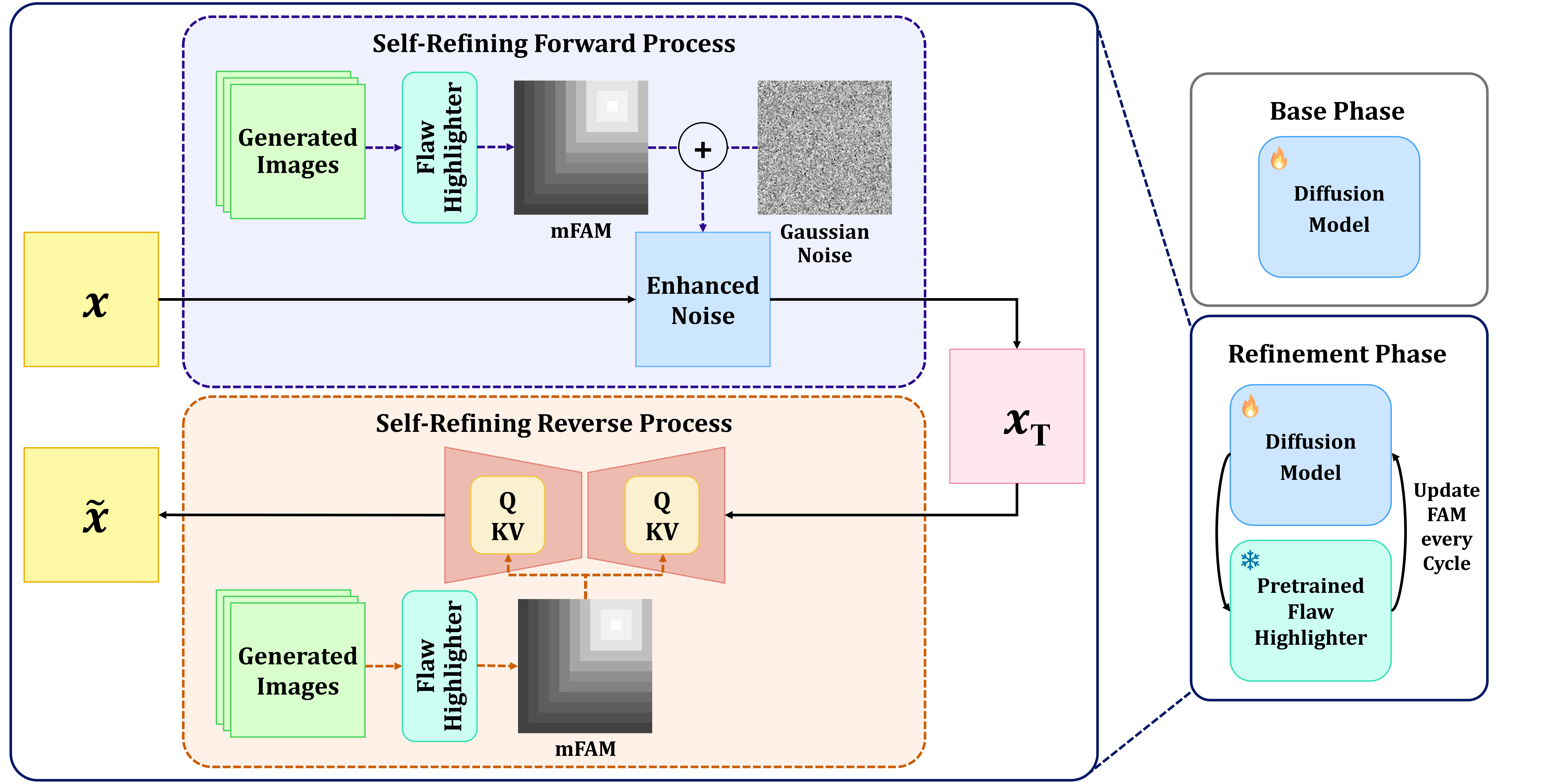}
  \caption{Training structure of the proposed self-refining diffusion framework. The proposed framework integrates flaw information from mean flaw attention map (mFAM) into both the forward and reverse processes, following a dual-phase training scheme. The base phase establishes baseline generation, and the refinement phase improves images quality using mFAM, with periodic updates ensuring continuous improvement.}
  \label{fig:training_structure}
\end{figure*}

Our contribution is summarized as follows:
\begin{itemize}
    \item We propose a novel framework, self-refining diffusion, that leverages a flaw highlighter to detect flaws in generated images and integrates this feedback into the training process, thereby improving the performance of diffusion models.
    \item Although prior studies have primarily focused on using XAI techniques for interpreting model behavior, we leverage them to actively guide the refinement of image generation.
    \item The proposed method is readily applicable to a variety of diffusion models and consistently outperforms standard approaches, improving image quality across various tasks and datasets.
\end{itemize}
\section{Related Work}
\label{sec:related_work}

\subsection{Diffusion Models}

Diffusion models have emerged as a highly effective class of generative models that progressively transform simple noise distributions into complex data distributions through iterative denoising \cite{yang2022survey}. Ho et al. \cite{ho2020ddpm} significantly improved this framework with their denoising diffusion probabilistic models (DDPMs), establishing a more stable training objective and demonstrating impressive results in various image generation tasks.

DDPMs \cite{ho2020ddpm} operate on a principled probabilistic framework consisting of two key processes. The forward process gradually adds Gaussian noise to data samples over T timesteps, transforming real data into pure noise. This process follows a Markov chain governed by a noise schedule $\beta_{1}, ..., \beta_{t}$, where each transition adds a small amount of noise to the data. The transition at each timestep is defined by a Gaussian distribution, as shown in \cref{eq:forward_process}.

\begin{equation}
  q(x_t \mid x_{t-1}) = \mathcal{N}(x_t; \sqrt{1 - \beta_t} \, x_{t-1}, \beta_t I)
  \label{eq:forward_process}
\end{equation}

The reverse process, which is learned during training, gradually recovers the original data by learning to predict the noise component at each timestep.

Building on the development of diffusion models \cite{rombach2022ldm,saharia2022photorealistic}, several studies have explored methods to guide the generation process and further improve output quality \cite{dhariwal2021beatgans,choi2021ilvr}. Unlike existing methods that enforce specific conditions or structural constraints, our approach uses an XAI-based flaw detection mechanism to directly identify and refine poorly generated regions.

\subsection{Explainable AI}

Explainable AI aims to make AI systems more interpretable by providing insights into their decision-making processes \cite{samek2019xai}. Initial work on XAI focused on developing post-hoc explanation methods for already trained models as well as designing inherently interpretable models \cite{moradi2021confident}.

\begin{figure*}[t]
  \centering
  \includegraphics[width=\textwidth]{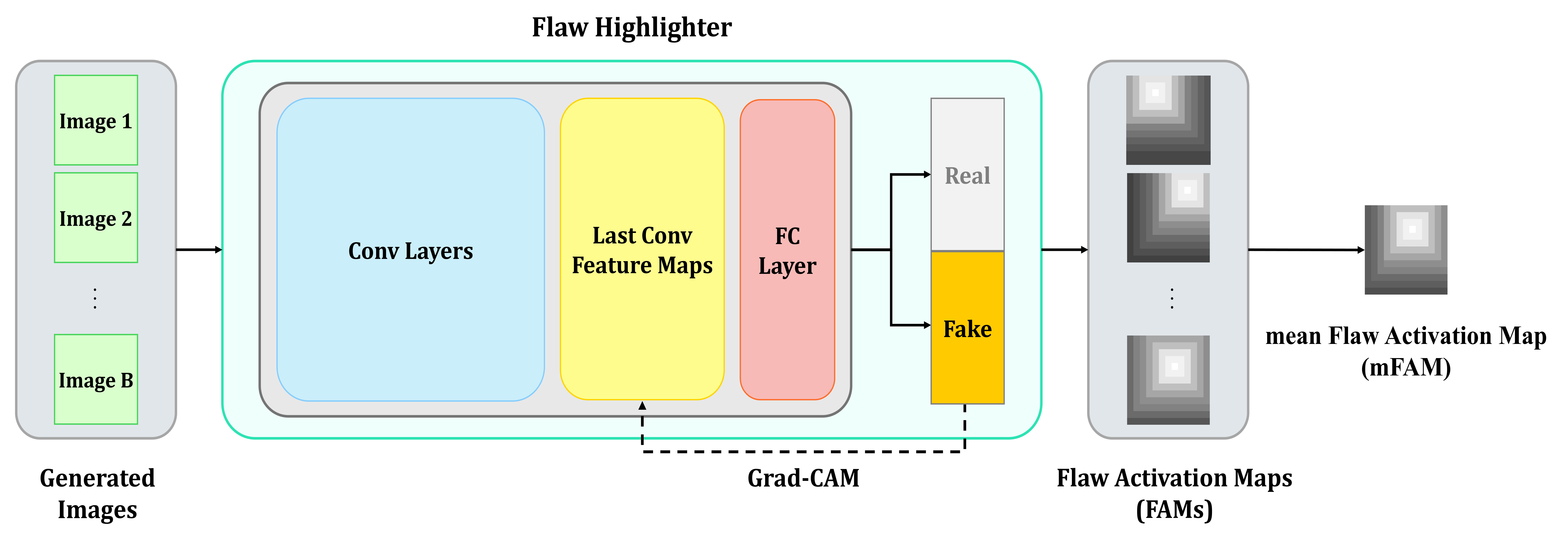}
  \caption{Architecture of the flaw highlighter. The flaw highlighter detects flawed regions in generated images using CAM, producing FAMs. The mFAM is obtained by averaging across the batch for more stable guidance.}
  \label{fig:flaw_highlighter}
\end{figure*}

CAM \cite{zhou2016cam} and its gradient-based extension, Grad-CAM \cite{selvaraju2017gradcam}, have emerged as powerful techniques for visualizing input regions that most influence a model’s decisions. Grad-CAM \cite{selvaraju2017gradcam} has been extensively adopted across diverse computer vision tasks, including image classification and object detection \cite{chattopadhyay2017gradcampp,fu2020axiom}.

Although XAI techniques have primarily been applied to classification tasks, their application to generative models presents unique challenges and opportunities. Bau et al. \cite{bau2020units} analyzed individual neurons in GANs to identify their semantic representations. Nagisetty et al. \cite{nagisetty2021xaigan} applied explainability techniques to identify key feature regions and incorporate them into the gradient updates of the generator. Tang et al. \cite{tang2023daam} introduced a method for visualizing cross-attention maps in diffusion models, enabling the interpretation of semantic alignment between text and images. These approaches primarily aim to explain model behavior rather than applying the obtained insights to improve generation quality. Our work transforms XAI mechanisms as active components in the refinement process of diffusion models, moving beyond their conventional role as diagnostic tools.
\section{Method}
\label{sec:method}

In this study, we propose a self-refining diffusion framework that improves the image generation quality of diffusion models by incorporating a flaw detection module, referred to as the flaw highlighter, grounded in XAI.

\subsection{Framework}

The proposed self-refining diffusion framework extends diffusion models by integrating a flaw highlighter to identify flawed regions within generated images. The identified flaws are incorporated into the forward and reverse processes of the diffusion model. To embed this mechanism effectively, we introduce a dual-phase training scheme coupled with a cycle-based update strategy, where saliency information is periodically refreshed. \cref{fig:training_structure} illustrates the overall training structure.

In the base phase, the diffusion model is trained following the standard diffusion model procedure without involvement from the flaw highlighter. This phase enables the model to learn core image generation capabilities, thereby establishing a solid foundation for subsequent flaw detection. In the refinement phase, while maintaining the core diffusion model training procedure, the model additionally incorporates defect information detected in the generated images. Specifically, FAMs obtained from a pretrained flaw highlighter are applied during both the self-refining forward and reverse processes, precisely guiding the denoising trajectory.

Rather than introducing a separate training stage, the proposed method replaces a portion of the standard training loop, allowing for efficient and seamless integration of the refinement mechanism. This integration enables continuous learning from the base phase to the refinement phase without disruption.

The flaw highlighter periodically updates the FAMs according to a predefined cycle schedule. This reflects the dynamic nature of flaw emergence as generative performance improves over training. Over successive training cycles, the self-refining diffusion framework progressively reduces artifacts in the generated images, ultimately producing results with higher visual fidelity and realism.

\begin{figure*}[t]
  \centering

  \setlength{\tabcolsep}{2pt} 
  
  \renewcommand{\arraystretch}{1.0} 

  \begin{tabular}{ccc}
    \textbf{CelebA-HQ (128$\times$128)} &
    \textbf{Oxford 102 Flower (128$\times$128)} &
    \textbf{LSUN Church (64$\times$64)} \\

    \includegraphics[width=0.31\linewidth]{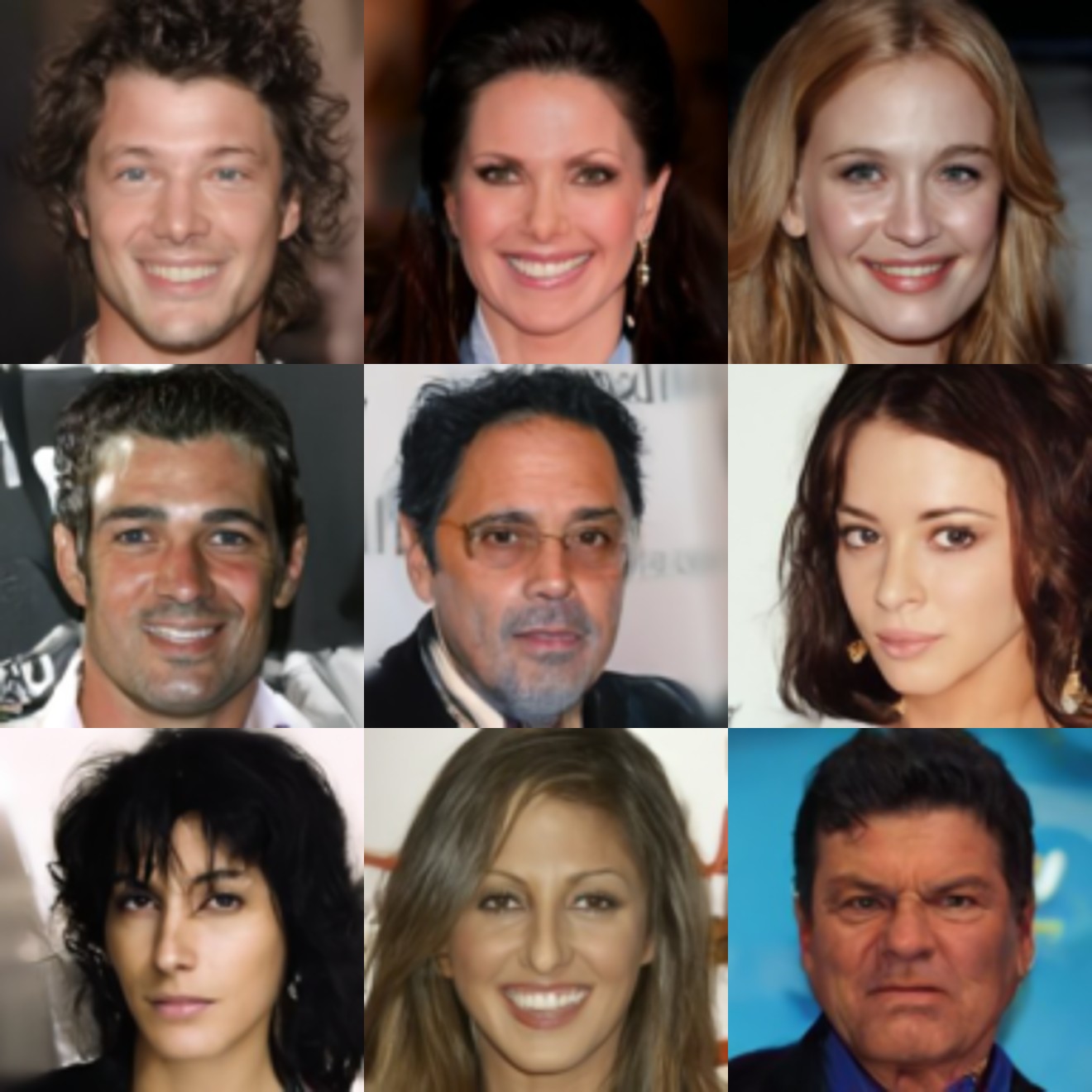} &
    \includegraphics[width=0.31\linewidth]{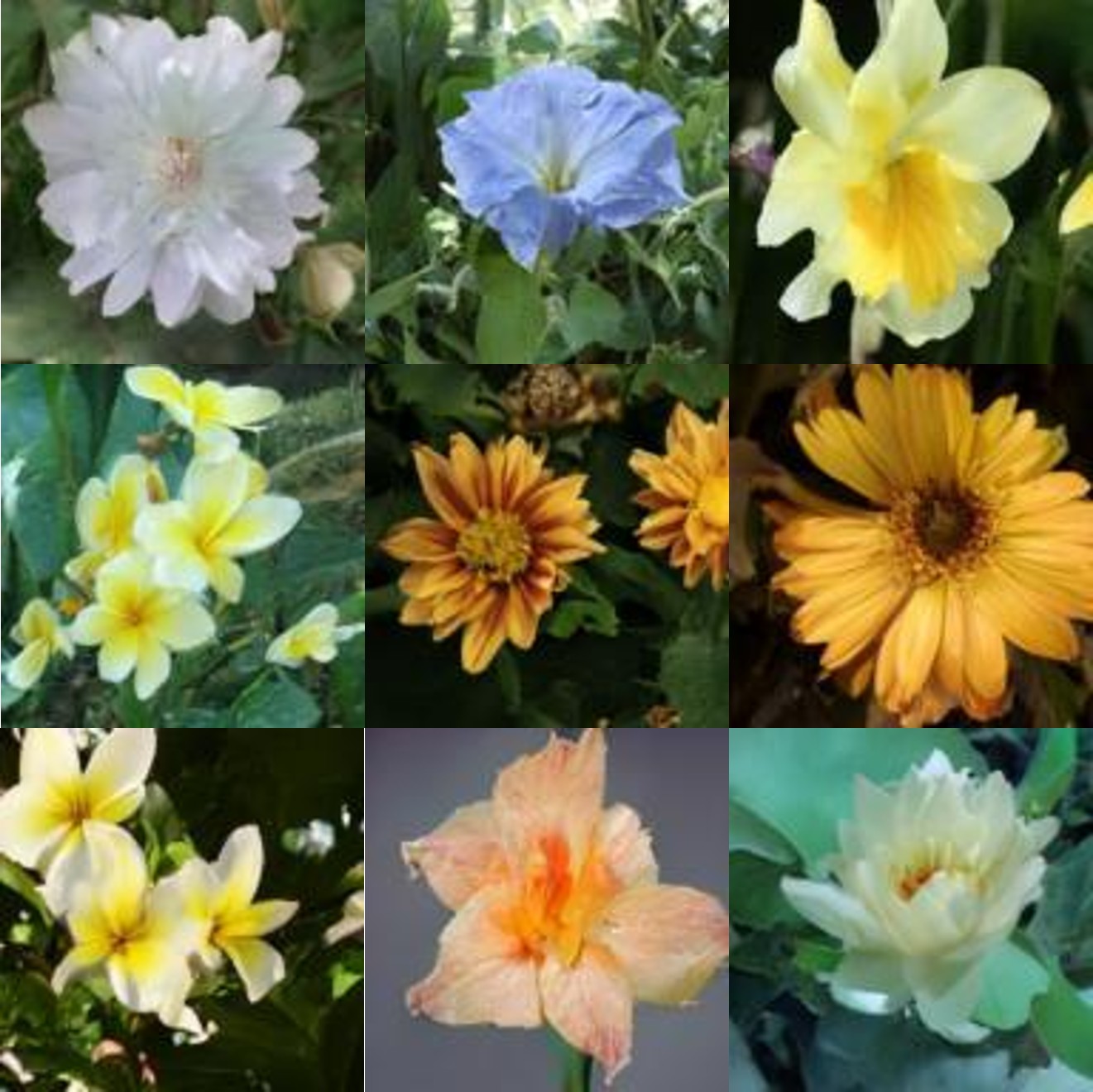} &
    \includegraphics[width=0.31\linewidth]{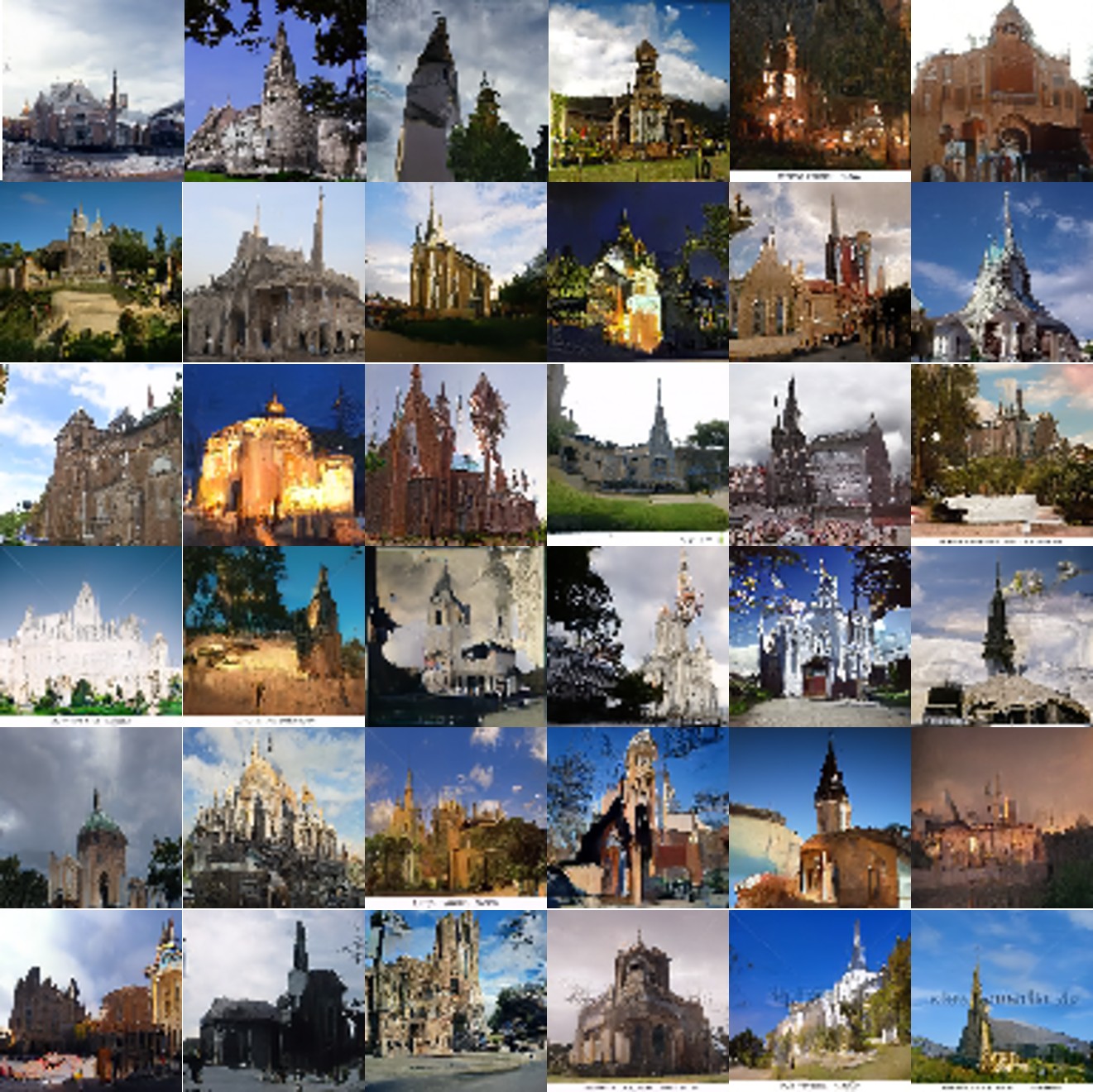} \\
  \end{tabular}
  
  \caption{Samples generated by self-refining diffusion from CelebA-HQ (128$\times$128), Oxford 102 Flower (128$\times$128), and LSUN Church (64$\times$64) datasets.}
  \label{fig:qualitative_samples}
\end{figure*}

\subsection{Flaw Highlighter}

The flaw highlighter is a model designed to detect unrealistic regions within images generated by a diffusion model, playing a critical role in enhancing image quality. As illustrated in \cref{fig:flaw_highlighter}, the proposed flaw highlighter is pretrained on a binary classification task that distinguishes between real images (real label) and generated images (fake label). To build the flaw highlighter, we employ a VGG16 backbone \cite{simonyan2015vgg}, which has demonstrated strong performance in prior image classification tasks. Once trained, the model receives a generated image as input and identifies regions that are likely to be unrealistic. Specifically, we apply CAM-based techniques (e.g., Grad-CAM) with respect to the fake class to generate FAMs that localize defective areas in the generated images. To further improve the robustness and consistency of the detected regions, we compute the mean flaw activation map (mFAM) by averaging FAMs across the batch.

\subsection{Self-Refining Forward Process}

As shown in \cref{fig:training_structure}, the self-refining forward process enhances the standard forward process of the diffusion model by integrating the mFAM into noise addition. This approach is based on the idea that injecting stronger noise into problematic regions improves reconstruction. Accordingly, we selectively enhance the noise level specifically in salient regions to increase the likelihood that the diffusion model will allocate greater attention to these areas during reconstruction. To achieve this, we combine the mFAM ($\bar{M}$) with the standard Gaussian noise used in diffusion models to produce an enhanced noise term, denoted as $\epsilon_{sr}$, which is then applied during the forward process. This modified noise injection is incorporated into the standard DDPMs \cite{ho2020ddpm} training loss, which is expressed as follows:

\begin{equation}
  \epsilon_{sr} = \epsilon + \lambda \overline{M}
  \label{eq:sr_noise}
\end{equation}

\begin{equation}
  \begin{gathered}
    \tilde{x}_t = \sqrt{\alpha_t} x_0 + \sqrt{1 - \alpha_t} \, \epsilon_{sr} \\
    L_{\text{simple}}(\theta) := \mathbb{E}_{x_0, \epsilon, t} \left[ \left\| \epsilon_{sr} - \epsilon_\theta(\tilde{x}_t, t) \right\|^2 \right]
  \end{gathered}
  \label{eq:loss_simple}
\end{equation}

In \cref{eq:sr_noise,eq:loss_simple}, $\epsilon$ represents the standard Gaussian noise employed in the diffusion model framework, and $\lambda$ is a modulation parameter that controls the extent of noise amplification based on the mFAM. The enhanced noise term $\epsilon_{sr}$ is defined as a selectively intensified version of $\epsilon$, where greater noise is injected into regions with higher mFAM values. Here, $x_0$ denotes the original clean image, and $\alpha_t$ is a noise schedule coefficient at timestep $t$. The timestep $t$ is sampled uniformly from $\{1, \cdots, T\}$ where $T$ is the total number of diffusion steps.

The proposed self-refining forward process encourages the model to focus on these salient areas, enabling finer reconstruction during denoising.

\subsection{Self-Refining Reverse Process}

As illustrated in \cref{fig:training_structure}, the self-refining reverse process introduces mFAM as a weighting factor within the attention mechanism of the diffusion model's denoising network. This mechanism is formally expressed through the following attention operation:
\begin{equation}
  K_{sr} = K \cdot (1 + \lambda \cdot \overline{M}_{\text{emb}}), \quad V_{sr} = V \cdot (1 + \lambda \cdot \overline{M}_{\text{emb}})
  \label{eq:attn_kv}
\end{equation}

Here, $K_{sr}$ and $V_{sr}$ denote the key and value matrices modulated by mFAM, derived from the original attention matrices $K$ and $V$, respectively. ${\bar{M}}_{emb}$ refers to the projected mFAM embedding. The parameter $\lambda$ controls the degree to which mFAM influences the attention mechanism. The embedding ${\bar{M}}_{emb}$ highlights semantically problematic regions within the image and is integrated into the attention mechanism to assign higher weights to the corresponding elements in the key and value matrices. As a result, regions with higher mFAM values receive more focused attention, improving the model’s ability to refine them during reconstruction.

\begin{figure*}[t]
  \centering
  \setlength{\tabcolsep}{2pt}

  \renewcommand{\arraystretch}{1.0}

  \begin{tabular}{ccc}
    \textbf{Self-Refining LDM (256$\times$256)} &
    \textbf{Self-Refining Improved DDPM (128$\times$128)} &
    \textbf{Self-Refining U-ViT (64$\times$64)} \\
    
    \includegraphics[width=0.31\linewidth]{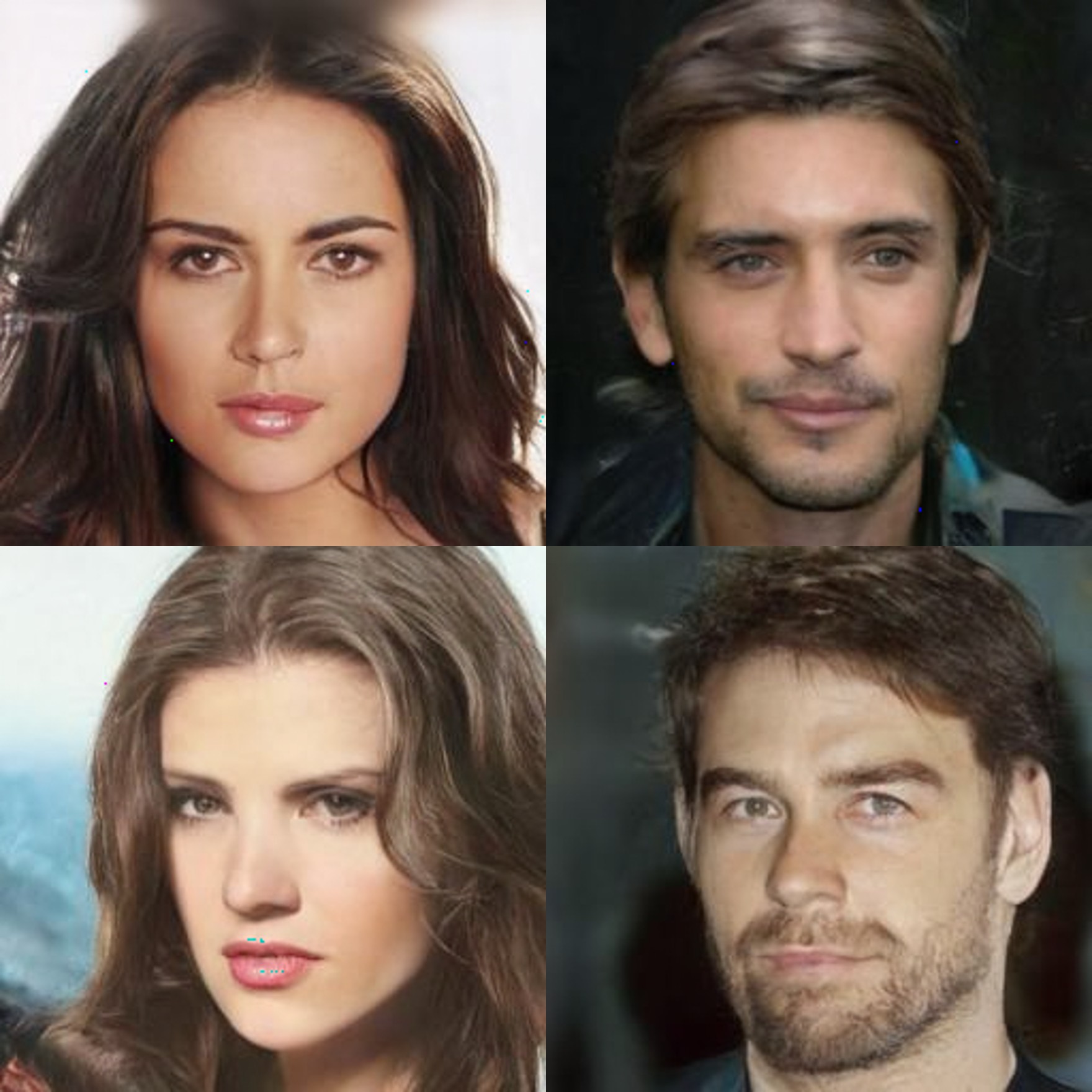} &
    \includegraphics[width=0.31\linewidth]{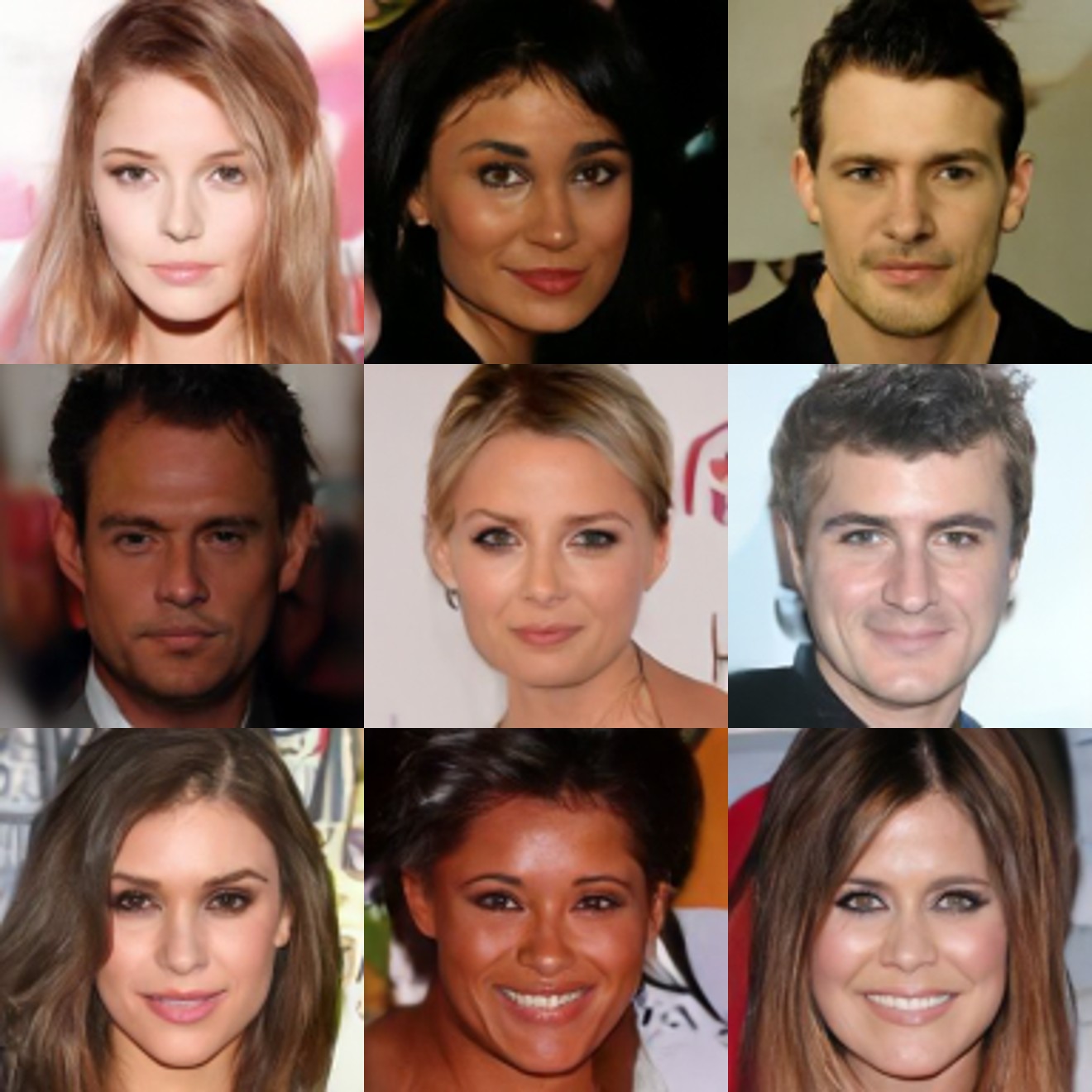} &
    \includegraphics[width=0.31\linewidth]{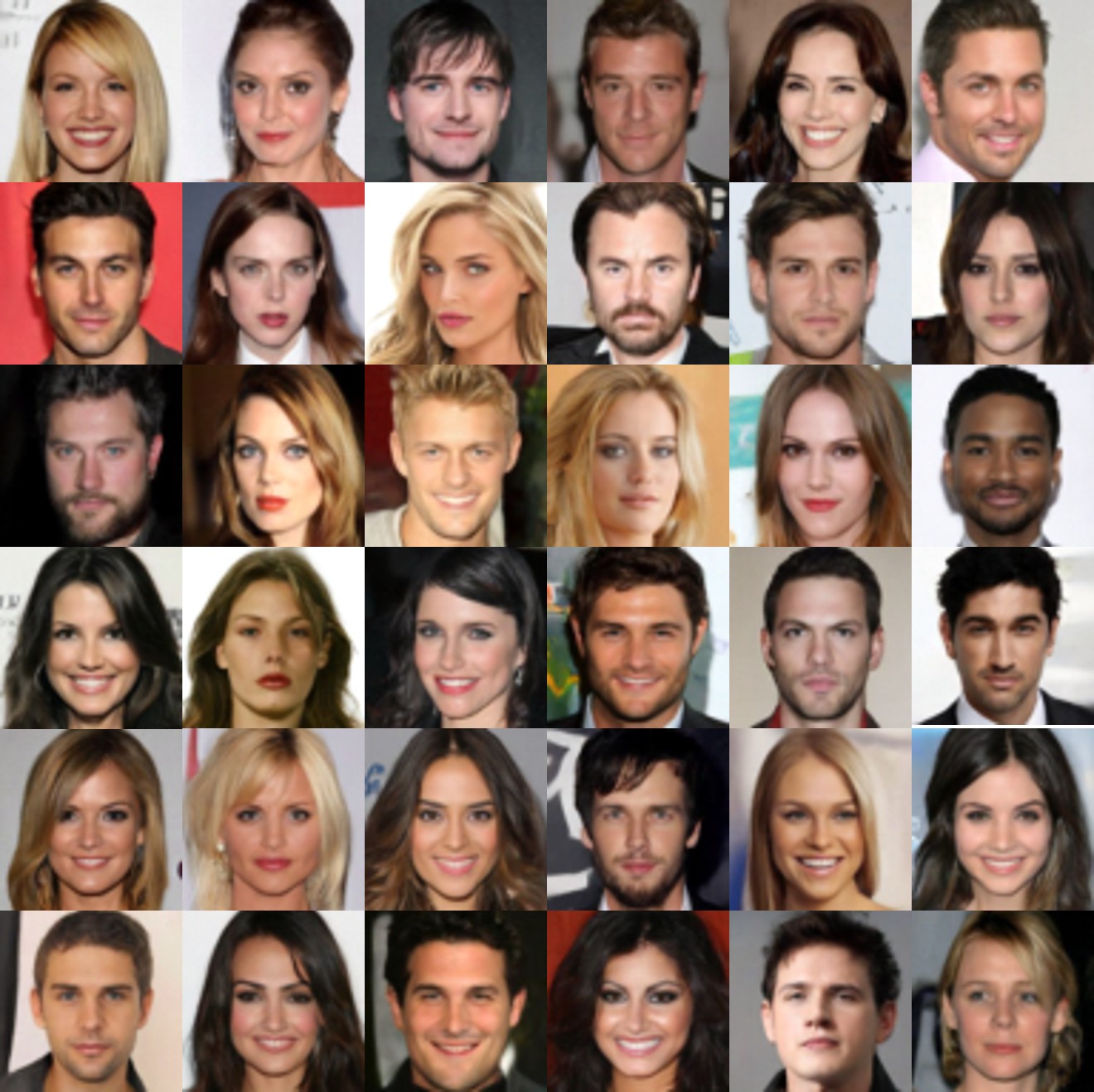} \\
  \end{tabular}

  \caption{Qualitative results of proposed self-refining framework applied to various diffusion models. From left to right: samples generated by self-refining LDM (256$\times$256), self-refining Improved DDPM (128$\times$128), and self-refining U-ViT (64$\times$64).}
  \label{fig:qualitative_models}
\end{figure*}

The self-refining reverse process follows the standard reverse diffusion procedure, augmenting it with mFAM-based attention modulation in the denoising network (e.g., U-Net and ViT) \cite{ronneberger2015unet, dosovitskiy2021an}. This addition enables more accurate reconstruction of flaw-prone areas, enhancing overall quality of generated images.

Overall, the proposed self-refining diffusion process leverages the flaw highlighter to achieve more accurate and consistent reconstruction of problematic image regions. Through these two refinement mechanisms, the diffusion model learns to better handle flawed regions than with conventional training. Once trained, the model performs inference exactly the same way as standard diffusion models, without requiring FAM. We discuss these improvements in detail in the following experimental section. 
\section{Experiments}
\label{sec:experiments}

In this section, the effectiveness of the proposed self-refining diffusion framework in image synthesis compared to fundamental diffusion models is evaluated. These model variants are referred to as “self-refining {model},” such as “self-refining DDPM.” The results of each variant are presented and compared in the following subsections. We conducted a performance comparison and analysis of the self-refining models, assessing their applicability across various tasks and the reliability of the proposed framework.

\subsection{Experimental Setup}

The proposed self-refining diffusion framework is compared with various diffusion-based models across multiple datasets. To assess the quality of generated images, we use FID \cite{heusel2017ttur} for image generation and employ learned perceptual image patch similarity (LPIPS) \cite{zhang2018perceptual}, structural similarity index measure (SSIM) \cite{wang2004ssim}, and peak signal-to-noise ratio (PSNR) \cite{nichol2021improvedddpm} for inpainting tasks.

\paragraph{Implementation Details} We run latent diffusion model (LDM) and Stable Diffusion \cite{rombach2022ldm} experiments on a single NVIDIA RTX 4090 GPU, while the remaining models and tasks are executed on a single NVIDIA RTX 4080 GPU. We conducted experiments by varying four hyperparameters: mFAM update cycles of 10, 50, 100, 500, and 1000 steps, phase ratios between the base and refinement phases of 1:4, 1:1, and 4:1, and mFAM weights in the forward and reverse processes across 0.01, 0.025, 0.05, and 0.1. The final configuration was determined based on validation performance.

\subsection{Self-Refining Diffusion for Image Generation}
\begin{table}[b]
  \centering

  \caption{Comparison of performance across different diffusion models and their self-refining variants using FID. (FID scores are reported as the mean $\pm$ standard deviation over five random seeds.)}
  \label{tab:fid_comparison}
  
  \begin{tabular}{lc}
    \toprule
    \textbf{Method} & \textbf{FID} $\downarrow$ \\
    \midrule
    DDPM \cite{ho2020ddpm} & 8.985 \\
    Self-Refining DDPM (ours) & \textbf{8.369 $\pm$ 0.044}\\ 
    \midrule
    Improved DDPM \cite{nichol2021improvedddpm} & 10.793\\
    Self-Refining Improved DDPM (ours) & \textbf{9.843 $\pm$ 0.064}\\
    \midrule
    U-ViT \cite{bao2023uvit} & 4.303\\
    Self-Refining U-ViT (ours) & \textbf{3.958 $\pm$ 0.049}\\
    \midrule
    LDM \cite{rombach2022ldm}& 9.269\\
    Self-Refining LDM (ours) & \textbf{8.116 $\pm$ 0.036}\\
    \bottomrule
  \end{tabular}
\end{table}

As shown in \cref{tab:fid_comparison}, the proposed self-refining framework consistently outperforms standard diffusion models on the CelebA-HQ dataset. In particular, our self-refining DDPM achieves a 6.9\% reduction in FID, indicating enhanced image quality. Similarly, the Improved DDPM with the proposed framework shows a 8.8\% reduction in FID. For U-ViT \cite{bao2023uvit}, the self-refining framework reduces the FID by 8\%, demonstrating its effectiveness even with ViT-based backbones. In the case of LDM \cite{rombach2022ldm}, the framework is applied only to the reverse process due to the model’s latent space structure, resulting in a 12.4\% reduction in FID. Overall, these results demonstrate that the proposed self-refining framework effectively improves the performance of various recent diffusion models, regardless of the backbone architecture, and operates effectively in latent space settings. Generated samples from self-refining DDPM are shown in Fig. \ref{fig:qualitative_samples}, and samples from other models are shown in Fig. \ref{fig:qualitative_models}.

\subsection{Comparison of Self-Refining Diffusion Processes}

We assess three model variants: the self-refining forward process, the self-refining reverse process, and the combined process, by evaluating our self-refining DDPM on the CelebA-HQ dataset (Karras et al. 2018) at 128 × 128 resolution. The training consists of 500K steps, with the base and refinement phases comprising 250K steps, respectively. In particular,  in the combined process variant, the update cycle of mFAM is set to 100, with the weights for the forward and reverse processes set to 0.01 and 0.025, respectively. As shown in  \cref{tab:ablation_process}, the combined application of forward and reverse processes achieves the most significant improvement. Notably, the reverse process outperforms the forward process, highlighting the effectiveness of incorporating saliency information into the attention mechanism of U-Net \cite{ronneberger2015unet}. This suggests that directing the focus toward semantically flawed regions during reconstruction enables more precise correction. Furthermore, the best results are obtained when both the forward and reverse processes are applied together, demonstrating the complementary benefits of noise amplification in the forward process and saliency-aware attention modulation in the reverse process.
\begin{table}[h]
  \centering
  
  \caption{Comparison of self-refining forward, reverse, and forward–reverse processes based on FID.}
  \label{tab:ablation_process}
  
  \begin{tabular}{lc}
    \toprule
    \textbf{Self-Refining Process} & \textbf{FID} $\downarrow$ \\
    \midrule
    Forward Process & 8.514 \\
    Reverse Process & 8.450 \\
    Forward-Reverse Process & \textbf{8.369} \\
    \bottomrule
  \end{tabular}
\end{table}

\subsection{Self-Refining Diffusion on Various Datasets}

To evaluate generalizability, we apply the proposed self-refining diffusion framework was evaluated on the Oxford 102 Flower \cite{nilsback2008flowers} and LSUN Church dataset \cite{yu2015lsun}. We train for 700K and 500K steps on the Oxford 102 Flower and LSUN Church datasets, respectively. For both datasets, the training is evenly divided between the base and refinement phases. \cref{tab:fid_extra_datasets} demonstrates that our framework consistently outperforms standard DDPM on both datasets, reducing FID by 27.3\% and 3.43\% on the Oxford 102 Flower and LSUN Church datasets, respectively. These results demonstrate the effectiveness and robustness of the proposed framework across diverse data domains. Sample outputs are shown in \cref{fig:qualitative_samples}.

\begin{table}[h]
  \centering
  \resizebox{\columnwidth}{!}{
  \begin{tabular}{@{}l c c@{}}
    \toprule
    \textbf{Method} & 
    \textbf{\begin{tabular}{@{}c@{}}Oxford Flower 102 \\ (FID) $\downarrow$\end{tabular}} & 
    \textbf{\begin{tabular}{@{}c@{}}LSUN Church \\ (FID) $\downarrow$\end{tabular}} \\
    \midrule
    DDPM \cite{ho2020ddpm} & 27.045 & 9.416 \\
    Self-Refining DDPM (ours) & \textbf{19.669} & \textbf{9.093} \\
    \bottomrule
  \end{tabular}
  }
  \caption{FID scores of DDPM and our self-refining DDPM (ours\_fr) on Oxford Flowers 102 (128$\times$128) and LSUN Church (64$\times$64).}
  \label{tab:fid_extra_datasets}
\end{table}

\subsection{Self-Refining Diffusion for Text-to-Image Generation}
Beyond image generation, we evaluate the proposed self-refining diffusion framework for text-to-image generation by integrating it into Stable Diffusion \cite{rombach2022ldm}, a widely used diffusion model for this task. Both the baseline Stable Diffusion and our self-refining variant are trained using the MS-COCO \cite{lin2014microsoft} dataset. Following the LDM, our framework is applied only to the reverse process due to the model’s latent-space formulation.
\begin{table}[h]
  \centering
  \caption{Quantitative comparison of text-to-image generation performance.}
  \label{tab:t2i_performance}
  
  \begin{tabular}{lc}
    \toprule
    \textbf{Method} & \textbf{FID} $\downarrow$ \\
    \midrule
    Stable Diffusion \cite{rombach2022ldm} & 25.571 \\
    Self-Refining Stable Diffusion (ours) & \textbf{24.633} \\
    \bottomrule
  \end{tabular}
\end{table}
\begin{figure}[b]
    \centering
    \rule[2pt]{\columnwidth}{0.4pt}
    \begin{tabular}{ccc}
        \parbox{0.3\columnwidth}{\centering\small\textit{"A zebra all by itself in the green forest."}} &
        \parbox{0.3\columnwidth}{\centering\small\textit{"A person is trying to ski down a snowy slope."}} &
        \parbox{0.3\columnwidth}{\centering\small\textit{"A plate is filled with broccoli and noodles."}} \\
    \end{tabular}
    \rule[2pt]{\columnwidth}{0.4pt}

    \vspace{0.5em}

    \begin{tabular}{ccc}
        \includegraphics[width=0.3\columnwidth]{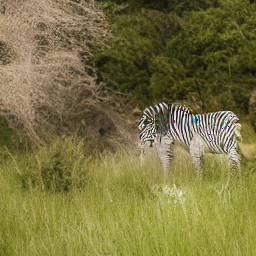} &
        \includegraphics[width=0.3\columnwidth]{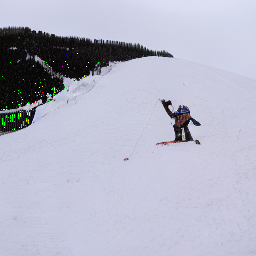} &
        \includegraphics[width=0.3\columnwidth]{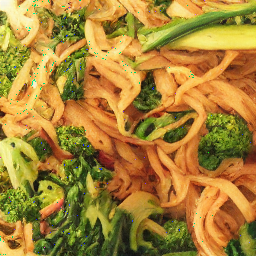} \\
    \end{tabular}

    \caption{Qualitative text-to-image examples generated by the proposed self-refining Stable Diffusion for several text prompts.}
    \label{fig:t2i_results}
\end{figure}
\begin{figure*}[t]
    \centering
    \setlength{\tabcolsep}{2pt}
    \renewcommand{\arraystretch}{1.0}
    \begin{tabular}{@{}c ccc@{}}
        \toprule
        & \textbf{Wide} & \textbf{Narrow} & \textbf{Altern. Lines} \\
        \midrule
        \textbf{GT} &
        \begin{tabular}{cc}
            \includegraphics[width=0.1\textwidth]{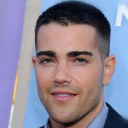} &
            \includegraphics[width=0.1\textwidth]{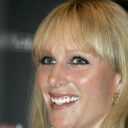}
        \end{tabular} &
        \begin{tabular}{cc}
            \includegraphics[width=0.1\textwidth]{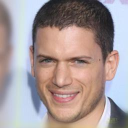} &
            \includegraphics[width=0.1\textwidth]{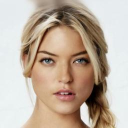}
        \end{tabular} &
        \begin{tabular}{cc}
            \includegraphics[width=0.1\textwidth]{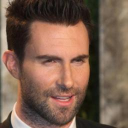} &
            \includegraphics[width=0.1\textwidth]{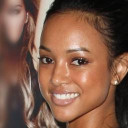}
        \end{tabular} \\
        \midrule
        \textbf{Masked} &
        \begin{tabular}{cc}
            \includegraphics[width=0.1\textwidth]{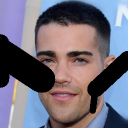} &
            \includegraphics[width=0.1\textwidth]{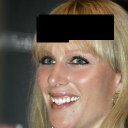}
        \end{tabular} &
        \begin{tabular}{cc}
            \includegraphics[width=0.1\textwidth]{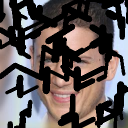} &
            \includegraphics[width=0.1\textwidth]{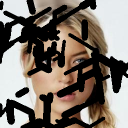}
        \end{tabular} &
        \begin{tabular}{cc}
            \includegraphics[width=0.1\textwidth]{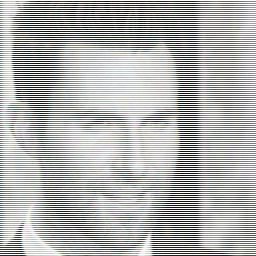} &
            \includegraphics[width=0.1\textwidth]{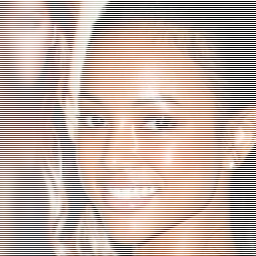}
        \end{tabular} \\
        \midrule
        \textbf{Result} &
        \begin{tabular}{cc}
            \includegraphics[width=0.1\textwidth]{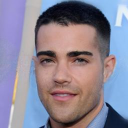} &
            \includegraphics[width=0.1\textwidth]{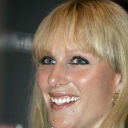}
        \end{tabular} &
        \begin{tabular}{cc}
            \includegraphics[width=0.1\textwidth]{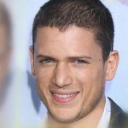} &
            \includegraphics[width=0.1\textwidth]{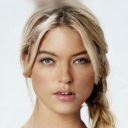}
        \end{tabular} &
        \begin{tabular}{cc}
            \includegraphics[width=0.1\textwidth]{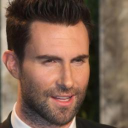} &
            \includegraphics[width=0.1\textwidth]{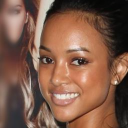}
        \end{tabular} \\
        \bottomrule
    \end{tabular}
    \caption{Inpainting results obtained using proposed self-refining diffusion framework. The figure compares the ground truth, masked inputs, and inpainted results under three conditions: wide, narrow, and alternating lines.}
    \label{fig:inpainting_results}
\end{figure*}
\begin{table*}[t]
  \centering
  \small
  \caption{Quantitative comparison of inpainting performance across three mask types (Wide, Narrow, and Alternating) using LPIPS, PSNR, SSIM, and FID.}
  \label{tab:inpainting_quantitative}
  \setlength{\tabcolsep}{2.3pt}
  \begin{tabular}{lcccccccccccc}
    \toprule
    \multirow{2}{*}{\textbf{Method}} & \multicolumn{4}{c}{\textbf{Wide}} & \multicolumn{4}{c}{\textbf{Narrow}} & \multicolumn{4}{c}{\textbf{Alternating Lines}} \\
    \cmidrule(lr){2-5} \cmidrule(lr){6-9} \cmidrule(lr){10-13}
     & FID↓ & PSNR↑ & SSIM↑ & LPIPS↓ & FID↓ & PSNR↑ & SSIM↑ & LPIPS↓ & FID↓ & PSNR↑ & SSIM↑ & LPIPS↓ \\
    \midrule
    RePaint \cite{lugmayr2022repaint} & 11.2213 & 23.3655 & 0.8356 & 0.0770 & 5.8492 & 30.0900 & 0.9292 & 0.0189 & 1.2314 & 49.6315 & \textbf{0.9973} & \textbf{0.0004} \\
    Self-Refining RePaint (ours) & \textbf{10.3564} & \textbf{23.3829} & \textbf{0.8365} & \textbf{0.0753} & \textbf{5.5396} & \textbf{30.0983} & \textbf{0.9305} & \textbf{0.0182} & \textbf{1.2211} & \textbf{49.6578} & \textbf{0.9973} & \textbf{0.0004} \\
    \bottomrule
  \end{tabular}
\end{table*}

As shown in \cref{tab:t2i_performance}, our method improves performance by 3.7\% over the baseline across standard evaluation metrics. Qualitative examples in \cref{fig:t2i_results} further demonstrate that our approach generates high-quality images that are well aligned with the input text. These results confirm the effectiveness of the proposed self-refining framework for this diffusion-based task.

\subsection{Self-Refining Diffusion for Inpainting}
We further evaluate our self-refining diffusion framework on image inpainting tasks. We adopt the Improved DDPM model \cite{nichol2021improvedddpm} as our base architecture and perform inpainting following the approach introduced in RePaint \cite{lugmayr2022repaint}. We use the CelebA-HQ dataset and LaMa-derived binary masks \cite{suvorov2022resolution}, including wide, narrow, and alternating lines types, each with 3,000 test images. As shown in \cref{tab:inpainting_quantitative}, the proposed method consistently outperforms the standard improved diffusion model across all evaluation metrics, including FID, PSNR, SSIM, and LPIPS. The inpainting results are presented in \cref{fig:inpainting_results}. This demonstrates that the proposed method, incorporating FAMs, excels visual perception and structural fidelity. This confirms the framework’s ability to generalize across tasks.

\subsection{Flaw Activation Map Analysis}

\paragraph{Flaw Highlighter Classification Performance} We evaluate the ability of the proposed flaw highlighter to distinguish between real and generated images using classification metrics and to localize artifacts via FAMs. \cref{tab:flaw_highlighter_performance} presents its classification performance on multiple datasets, where the generated images are produced by diffusion models. Across all three datasets, the proposed flaw highlighter consistently exhibits high classification performance, reliably distinguishing real and generated images. Notably, on the CelebA-HQ dataset, the proposed flaw highlighter demonstrates outstanding performance across all three metrics.

\begin{table}[h]
  \centering
  \caption{Performance of proposed flaw Highlighter across different datasets. The accuracy, F1-score, and ROC-AUC are reported.}
  \label{tab:flaw_highlighter_performance}
  \small
  \resizebox{\columnwidth}{!}{
    \begin{tabular}{lccc}
      \toprule
      \textbf{Dataset} & \textbf{Accuracy ↑} & \textbf{F1 Score ↑} & \textbf{ROC AUC ↑} \\
      \midrule
      CelebA-HQ (64$\times$64) & 0.794 & 0.795 & 0.875 \\
      CelebA-HQ (128$\times$128) & 0.991 & 0.991 & 1.000 \\
      CelebA-HQ (256$\times$256) & 0.990 & 0.990 & 0.999 \\
      Oxford 102 Flower & 0.882 & 0.880 & 0.993 \\
      LSUN-Churches & 0.900 & 0.900 & 0.966 \\
      \bottomrule
    \end{tabular}
  }
\end{table}

\paragraph{Comparison of Guidance Maps}
To validate our FAM, we compared it with three alternative guidance maps, all applied within the same self-refining framework. The baselines were as follows: (1) Center-Gaussian Map — a static map emphasizing the image center, with a value of 1 at the center and 0 at the edges; (2) Inverted-Gaussian Map — its inversion, emphasizing the periphery with 0 at the center and 1 at the edges; and (3) Edge Map — a structural map derived from the OpenCV Canny \cite{canny2009computational} operator, emphasizing high-frequency details. \cref{tab:fid_guidance} presents the FID comparison for these maps on the CelebA-HQ (128×128) dataset, with DDPM as the base model.

\begin{table}[h]
  \centering
  \caption{Comparison of FID scores for different guidance maps.}
  \label{tab:fid_guidance}

  \begin{tabular}{lc}
    \toprule
    \textbf{Guidance Map} & \textbf{FID $\downarrow$} \\
    \midrule
    Center-Gaussian Map & 10.103 \\
    Inverted-Gaussian Map & 10.963 \\
    Edge Map & 8.739 \\
    Flaw Activation Map (FAM) & \textbf{8.369} \\
    \bottomrule
  \end{tabular}
\end{table}

The static, spatially-biased maps (Center-Gaussian Map, Inverted-Gaussian Map) yielded the worst performance, proving a simple, fixed focus is ineffective. While the structural Edge Map performed significantly better, the proposed FAM achieved the best results. This strongly suggests that a learned map that dynamically identifies semantically flawed and unrealistic regions provides a more effective and targeted guidance signal for refinement than either static spatial biases or purely structural maps.
\begin{figure}[b]
    \centering
    \setlength{\tabcolsep}{1pt}

    \begin{tabular}{@{}c cccc@{}}
        \makecell[c]{\scriptsize\textbf{Generated} \\ \scriptsize\textbf{Image}} &
        \raisebox{-0.5\height}{\includegraphics[width=0.2\columnwidth]{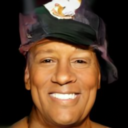}} &
        \raisebox{-0.5\height}{\includegraphics[width=0.2\columnwidth]{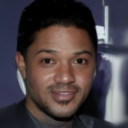}} &
        \raisebox{-0.5\height}{\includegraphics[width=0.2\columnwidth]{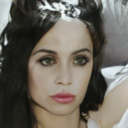}} &
        \raisebox{-0.5\height}{\includegraphics[width=0.2\columnwidth]{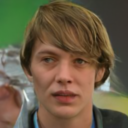}} \\
        
        \addlinespace[2pt]

        \makecell[c]{\scriptsize\textbf{Heatmap}} &
        \raisebox{-0.5\height}{\includegraphics[width=0.2\columnwidth]{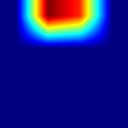}} &
        \raisebox{-0.5\height}{\includegraphics[width=0.2\columnwidth]{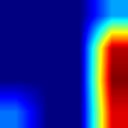}} &
        \raisebox{-0.5\height}{\includegraphics[width=0.2\columnwidth]{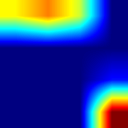}} &
        \raisebox{-0.5\height}{\includegraphics[width=0.2\columnwidth]{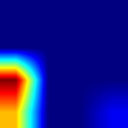}} \\

        \addlinespace[2pt]

        \makecell[c]{\scriptsize\textbf{Overlay}} &
        \raisebox{-0.5\height}{\includegraphics[width=0.2\columnwidth]{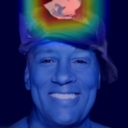}} &
        \raisebox{-0.5\height}{\includegraphics[width=0.2\columnwidth]{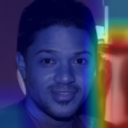}} &
        \raisebox{-0.5\height}{\includegraphics[width=0.2\columnwidth]{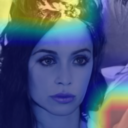}} &
        \raisebox{-0.5\height}{\includegraphics[width=0.2\columnwidth]{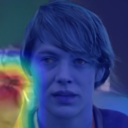}} \\
    \end{tabular}
    
    \caption{Standard DDPM-generated images, FAM heatmaps, and overlay results, accurately highlighting defect-prone regions.}
    \label{fig:fam_visualization}
\end{figure}

\paragraph{Visualization of Flaw Activation Maps} \cref{fig:fam_visualization} illustrates how the FAMs accurately identify the most unrealistic regions in the DDPM-generated images. In the generated outputs, subtle artifacts remain around the eyes and eyebrows, the hairline boundaries appear blurred, and irregular noise is visible around the head and in the background. The FAM heatmaps assign the highest activations (warm red and yellow) to areas including the left eyebrow region and hair-background boundary. When these maps are overlaid on the generated images, the FAM hotspots precisely align with actual defect-prone spots, confirming that the proposed flaw highlighter effectively captures the flaw patterns of the model and provides reliable refinement guidance.

\begin{figure}[h]
    \centering
    \setlength{\tabcolsep}{1pt}

    \begin{tabular}{@{}c cccc@{}}
        \makecell[c]{\scriptsize\textbf{Generated} \\ \scriptsize\textbf{Image}} &
        \raisebox{-0.5\height}{\includegraphics[width=0.2\columnwidth]{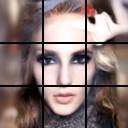}} &
        \raisebox{-0.5\height}{\includegraphics[width=0.2\columnwidth]{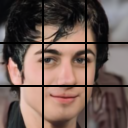}} &
        \raisebox{-0.5\height}{\includegraphics[width=0.2\columnwidth]{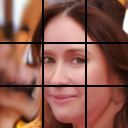}} &
        \raisebox{-0.5\height}{\includegraphics[width=0.2\columnwidth]{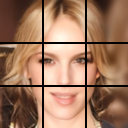}} \\
        
        \addlinespace[2pt]

        \makecell[c]{\scriptsize\textbf{Overlay}} &
        \raisebox{-0.5\height}{\includegraphics[width=0.2\columnwidth]{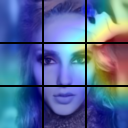}} &
        \raisebox{-0.5\height}{\includegraphics[width=0.2\columnwidth]{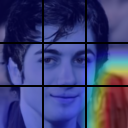}} &
        \raisebox{-0.5\height}{\includegraphics[width=0.2\columnwidth]{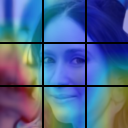}} &
        \raisebox{-0.5\height}{\includegraphics[width=0.2\columnwidth]{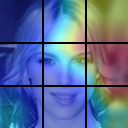}} \\
    \end{tabular}
    
    \caption{Comparison of human-identified artifact regions and FAM overlays. The figure shows DDPM-generated images (top) and FAM overlays (bottom), with highlighted regions aligning with human-perceived flaws.}
    \label{fig:human_comparison}
\end{figure}

\paragraph{Human Perceptual Evaluation of Flaw Activation Map} As shown in \cref{fig:human_comparison}, we conducted a user study to evaluate whether our FAMs correspond to the regions that humans perceive as artifacts. Fifteen participants were shown a grid-tiled DDPM-generated face image and selected the three grid cells they perceived as most unnatural. For each image, the three cells highlighted by human observers were compared with the top three highest-activation cells in the FAM. If a human-selected cell coincided with one of the FAM’s top three cells, it was counted as a match. Across all images and participants, we observed an average match rate of 87\%, indicating that the FAM hotspots aligned closely with the regions the human viewers identified as flawed. The strong agreement confirms that the proposed flaw highlighter not only performs well on quantitative classification metrics but also captures the very artifacts that are visually salient to people, highlighting its value as a tool for guiding targeted refinement.

\section{Conclusion}
\label{sec:conclusion}

In this study, we proposed self-refining diffusion, a novel generative framework that leverages XAI techniques to detect and correct flawed regions in generated images. By utilizing FAMs, the proposed method highlights artifact-prone areas and integrates this information into both the forward and reverse diffusion processes, thereby improving the performance of diffusion models.

Quantitative evaluations demonstrate that self-refining diffusion significantly outperforms the standard baselines of various diffusion-based models across diverse tasks and datasets. Furthermore, qualitative user studies demonstrate that FAMs effectively capture visual flaws that align with human perception, enhancing both performance and interpretability. These results confirm that our framework provides a versatile and robust approach, demonstrating significant improvements across a wide range of diffusion models and generative tasks. 
{
    \small
    \bibliographystyle{ieeenat_fullname}
    \bibliography{main}
}

\end{document}